\documentclass[letterpaper, 10 pt, conference]{ieeeconf}  
\usepackage{amsmath}
\IEEEoverridecommandlockouts                              
\usepackage{graphicx}                                                
\usepackage{todonotes}                                                          
\title{\LARGE \bf
The Analysis of Discrete-Event System in Autonomous Package Delivery using Legged Robot and Conveyor Belt
}

\author{Garen Haddeler$^{*}$
\thanks{ $^{*}$Department of Mechanical Engineering, National University of Singapore, 117575, Singapore, {\tt\small 	e0444217@u.nus.edu} }
}

\begin{document}

\maketitle
\thispagestyle{empty}
\pagestyle{empty}

\begin{abstract}
In this paper, the supervisory control of a Discrete Event System (DES) analyses states and events to construct autonomous package delivery system. 
The delivery system includes legged robot in order to autonomously navigate uneven indoor terrain and a conveyor belt for transporting the package to  the legged robot. 
The aim of the paper  is using  theory  of supervisory  control  of DES  to supervise and control  machine’s state and event and  ensure robots autonomously collaborate.
By applying the theory, we show collaboration of two individual robots to deliver goods in multi-floor environment
The obtained results from the theory of supervisory control is implemented and verified in simulation environment.
\end{abstract}

\section{INTRODUCTION}
Delivering package in indoor and uneven terrain can be challenging since today's robot cannot fully represent and navigate multi-storey terrain. Comparing with wheeled robots, legged robots can be used to navigate uneven terrains since legged robots can overcome larger obstacles than their body frame \cite{bosworth_kim_hogan_2015}. 
Inspired by the capability of such robots, we developed an autonomous navigation framework for the legged robot to fulfil desired behaviour which for our case reaching the goal, avoiding obstacles, climbing stairs and deliver goods. Moreover, a conveyor belt is used to initially transfer goods to the robot's body frame. Thus, robot and conveyor need to autonomously communicate with each other and perform delivery. To do so, we developed a supervisory control strategy to control individually machine’s states and events and ensure robot and conveyor autonomously work together safely and efficiently. The obtained supervisory control strategy is verified on the simulation environment and legged navigation framework. The code is available in the \textbf{link \footnote{https://github.com/hgaren/aliengo$\_$delivery}}.

There are several autonomous robots project were used for package deliveries. Amazon's warehouses are actively using autonomous swarm robots which they deliver goods between one point to another using conveyor belts and wheeled robots \cite{swarm_robot}. However, the main drawback of wheeled vehicles are goods that can not deliver in multi-storey facilities. To overcome this issue legged robot can be proposed instead of wheeled. 
A bipedal robot from Agility Robotics is used to deliver goods in multi-storey outdoor terrain \cite{ford_robot}. Similarly, Boston Dynamic's wheeled-bipedal robot named Handle can perform delivery in indoor facilities \cite{bjelonic_bellicoso_viragh_sako_tresoldi_jenelten_hutter_2019}. However, their approach doesn't include collaboration with a conveyor belt and multiple robot aspect. By using Supervisory control theory for DES system, we propose autonomous collaboration between the legged robot and conveyor belt.
 
The following parts of this paper are arranged as follows: Section~\ref{sec:preliminaries} introduces the general concept of autonomous legged robot navigation and supervisory control theory. Thereafter, subsection ~\ref{sec:automation}  shows the autonomous and system structure of a task-based delivery scenario. Moreover, the proposed supervisory controlled structure is verified through simulation in Section~\ref{sec:results}. In the end, Section~\ref{sec:conclude} concludes the work and gives an outlook on future research. 

\section{Preliminaries}
\label{sec:preliminaries}
In this section, concept of legged robot and its autonomous navigation are presented. 
Moreover, some core methodologies of supervisory control and automata are introduced. These methodologies are used to construct our package delivery scenario.
\subsection{Legged Robot}
Autonomous navigation is needed to deliver goods from its starting position to the desired goal position.  Generally, in legged robots, self-driving behaviour can be carried out three sub-modules: robot localization, path planning and body controller.
\subsubsection{Robot localization}
Localization is one of the major steps to perform autonomous
navigation, path tracking, avoiding obstacles and mapping. Therefore, in this article, ready localization algorithm
is used for navigation purposes.
Since this isn't the main focus of this work, we obtained the robot's pose from the simulation environment.

\subsubsection{Path Planning}
Path planning algorithm is used to generate the path in between starting to goal positions and avoid static obstacles. 
We used 2D NavfnROS path planning algorithm to plan path at the pre-defined route.

\subsubsection{Body Controller}
Comparing with wheeled robots, legged robot's body controller has higher complexity due to its dynamics and kinematics. One of the main reason is legged robots can be unstable while moving and needs to have a robust controller to balance itself.
We used MIT cheetah's body controller to balance its body frame and perform foot placement from the gait generator according to command reference (velocity).
Note that the legged robot needs to climb stairs therefore we use local awareness system to re-plan foot placements if the location of the planned foot is not traversable \cite{bosworth_kim_hogan_2015}. 
Additionally, local awareness' perception is obtained by traversability grid map where we evaluate terrain's slope and roughness information to detect step-able areas \cite{wermelinger_fankhauser_diethelm_krusi_siegwart_hutter_2016}.
\subsection{Concepts of automata and supervisory control}
The supervisory control theory allows us to observe free events which are all possible combinations of events, and control controllable events so that agents fulfil given specifications \cite{control_des}.

Automata which is represented by Discrete event-states (DES) consists of a set of states and a set of events.
An event causes a DES system to move from one state to another state and these events are assumed to occur instantaneously.
Below formulation where states,$Q$,  are represented by numbers and events $\sum$, are represented by symbols as following
\begin{equation}
    Q = \begin{Bmatrix}
0 & 1 &2
\end{Bmatrix} \text{  }
    \sum =\begin{Bmatrix}
 \alpha & \beta & \gamma 
\end{Bmatrix} 
\end{equation}
 Automaton is constructed
and represented as follow eq. \ref{eq:automaton} where $Q$ is states, $\sum$ is events, $\delta$ transition function (which is relationship between current state, event and new state),  $q_0$ indicates initial state and lastly, $Q_m$  is marked states (which  usually refers to system's equilibrium states).
\begin{equation}
\label{eq:automaton}
   G = \begin{Bmatrix}
 Q & \sum & \delta & q_0 & Q_m 
\end{Bmatrix}
\end{equation}
In order to present a sequence of the events, a string is
defined as the sequence of symbols which may contain multiple
symbols or one symbol. Language refers to the collection and combination of
event and generated from automaton. An example of language and marked language can be shown as below eqs respectively \ref{eq:language}, \ref{eq:language_marked} . Note that marked language needs to satisfy $L_m(G) \subseteq L(G)$.
\begin{equation}
\label{eq:language}
   L(G) =\begin{Bmatrix}
 \epsilon & \alpha \beta  &  \alpha  \beta  \gamma  & \alpha  \beta \alpha \gamma &  ...
\end{Bmatrix}
\end{equation}
\begin{equation}
\label{eq:language_marked}
   L_m(G) =\begin{Bmatrix}
 \alpha \beta  &  \alpha  \beta  \gamma  ...
\end{Bmatrix}
\end{equation}
The supervisory control theory defines a supervisor function $S$ which also called control policy so that it can control all controllable events of plant  $G$ where it follows given specification $E$. Controlled behavior represented as $S/G$ and note that marked languages need  to satisfy $L_m(S/G) \subseteq L_m(E)$ \cite{control_des}.

Briefly, by obtaining free behaviour of plant $G$ and defined specifications  $E$, we can extract supervisor function $S$ such that it can generate controlled DES and it's language $L_m(S/G)$. Further section \ref{sec:automation}, we will show how we obtained our scenario's free behaviour and it's the specification.

\section{Automation and System Structure}
\label{sec:automation}
Our system has initially two real machines one robot and one conveyor.  
The delivery scenario starts with a legged robot spawning on the first floor. Secondly, the robot’s first state is initialized by the operator and it navigates through the first goal which is in front of the conveyor belt. 
Next, robot docks to belt and conveyor start moving to transfer the package to the robot’s designated rectangular area.
After conveyor belt finishes moving to place package, the robot stands up.
Then, by climbing stair, the robot tries to reach the second goal where is on one upper floor. 
Lastly, if the robot reaches its second goal, the robot changes its state to success state and finishes its task until operator resends the first goal.

To resolve task complexity infinite state machine wise, we divided our legged robot into two machines, therefore, our whole system consists of three sub-machines: First Task Robot, Conveyor Belt, Second Task Robot.
\subsection{Machine-1: First Task Robot}
\begin{figure}
     \includegraphics[width=0.3\textwidth]{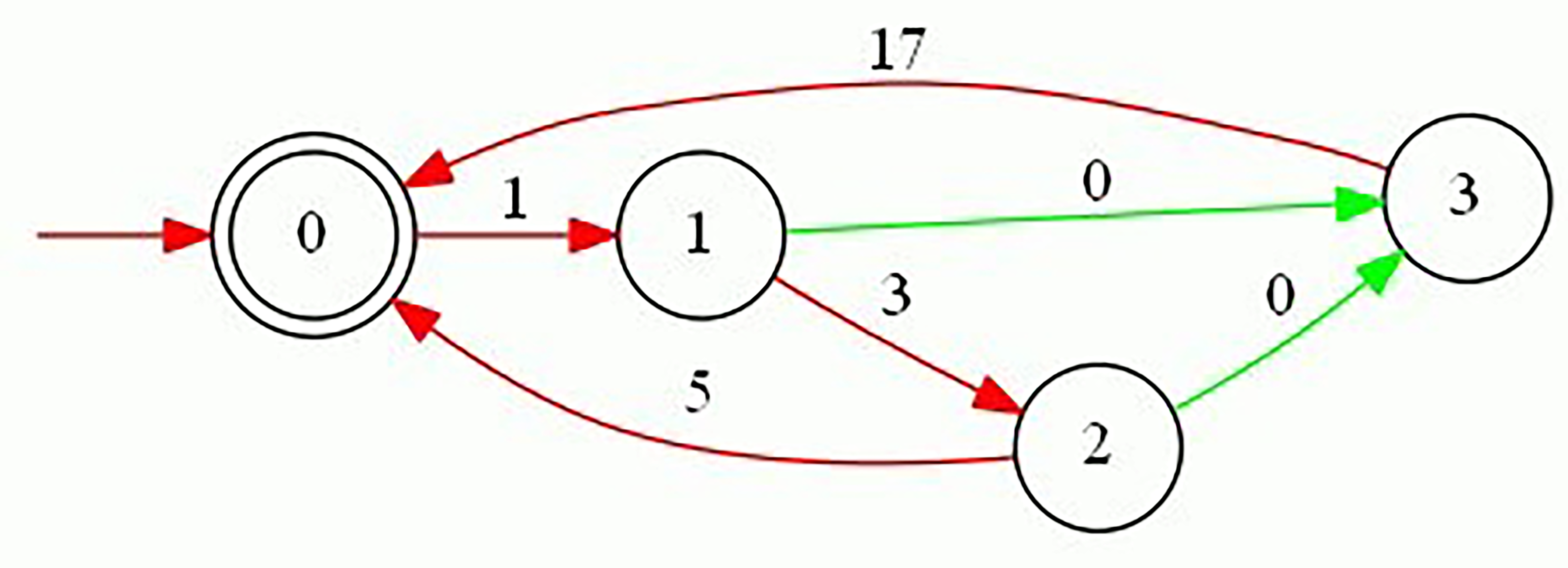}
     \caption{ The automaton model of Machine-1 First Task Robot: (Red) controllable, (Green) uncontrollable events}
     \label{fig:machine_1}
 \end{figure}

Machine-1 is defined as reaching the first goal using the legged robot. Figure \ref{fig:machine_1} shows states and events of the system using Automata. Table \ref{table:machine_1} defines states-events and their representations. 

According to our delivery scenario, initially, the robot is in the idle state which means the robot is in stable-stopping position. An operator starts the delivery scenario by giving the first goal to the robot. Thereafter walking state starts and the legged robot uses the proposed autonomous navigation framework to reach its first goal. After reaching its first goal, the robot changes its state to docking and robot docks through the conveyor belt. In this state, robot slowly approaches to belt and descends to collect the package.  After docking finished robot reaches idle state again and wait for the conveyor belt to transfer the package. As an uncontrollable event, if the robot fails while walking or docking change its state to fail and by an error flag event robot's state becomes idle.
 \begin{table}
 \caption{Table of State and Events: Machine-1 First Task Robot}
\label{table:machine_1}
\begin{tabular}{ |c|c|c| } 
 \hline
 States & Events  \\ 
  \hline
 Robot Idle:0  &  First goal started:1  \\ 
 Walking:1  & First goal reached:3, Robot failed:0\\ 
  Docking:2  & Docking finished:5, Robot failed:0\\ 
  Fail:3  & Error flag:17 \\
 \hline
\end{tabular}
\end{table}

\subsection{Machine-2: Conveyor Belt}
\begin{figure}
     \includegraphics[width=0.3\textwidth]{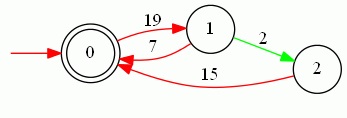}
     \caption{The automaton model of  Machine-2 First Task Robot: (Red) controllable, (Green) uncontrollable events}
     \label{fig:machine_2}
 \end{figure}

Machine 2 is defined as conveyor belt transferring goods to a robot. Figure \ref{fig:machine_2} shows states and events of the system using automata. Table \ref{table:machine_2} defines states-events and their representations. 

According to our delivery scenario, the conveyor belt is idle state until the robot docking is finished. Thereafter, conveyor belt changes its state to a working state and package is moved to robot's top. After package moving finished conveyor changes its state to idle again. As an uncontrollable event, if conveyor belt drops a package to the ground in a working state, it changes its state to fail state. Thereafter, box re-spawn on the conveyor belt, machine-2 change it's stated to idle and the whole process can start again.
 
 \begin{table}
 \caption{Table of State and Events: Machine-2 Conveyor Belt}
\begin{tabular}{ |c|c|c| } 
 \hline
 States & Events  \\ 
  \hline
 Conveyor Idle:0  &  Moving Box:19  \\ 
 Walking:1  & Stopping box:7, Box dropped:2\\ 
  Fail:2  &  Spawn box: 15 \\
 \hline
\end{tabular}
\label{table:machine_2}
\end{table}
\subsection{Machine-3: Second Task Robot}
\begin{figure}
     \includegraphics[width=0.3\textwidth]{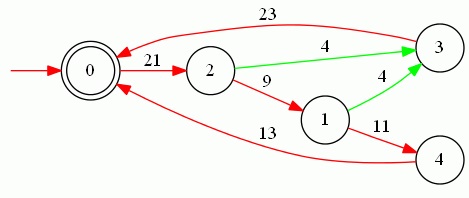}
     \caption{The automaton model of Machine-3 First Task Robot: (Red) controllable, (Green) uncontrollable events}
     \label{fig:machine_3}
 \end{figure}

Machine-3 is defined as reaching second goal using legged robot. Comparing with machine-1, identical events and states represent different values. Figure  \ref{fig:machine_3}  shows states and events of the system using automata. Table \ref{table:machine_3} defines states-events and their representations.

Similarly, with Machine-1, initially, the robot is in the idle state which means the robot is in stable-stopping position.Machine-3 start the stand-up state after the package is on the robot's base. Thereafter walking state starts and the legged robot uses the proposed autonomous navigation framework to reach its second goal where is one level above from pick-up position. After reaching its second goal, the robot changes its state to success and later, by a success flag, it changes to an idle state to finish it's a task and stop until operator gives the first goal again. As an uncontrollable event, similarly with Machine-1, if the robot fails while walking or standing up change its state to fail and by an error flag event, the robot's state becomes idle
 
 \begin{table}
 \caption{Table of State and Events: Machine-3 Second Task Robot}
\label{table:machine_3}
\begin{tabular}{ |c|c|c| } 
 \hline
 States & Events  \\ 
  \hline
 Robot Idle:0  & Box is on the robot:21  \\ 
 Walking:1  & Second goal reached:11,
Robot failed:3
\\ 
  Stand up:2  & Second goal started:9,
Robot failed:3\\ 
  Fail:3  & Error flag:23 \\
    Success:4  & Success flag:13 \\

 \hline
\end{tabular}
\end{table}
\subsection{Specifications}
According to Supervisory theory, specifications are needed to control free behavior of the system. To do so, we have defined 8 specification as follows:
	\begin{enumerate} 
     \item After docking finished (5), moving box (19) event must start.
     \item After moving box (19), stopping box (7) event must start or as an uncontrollable event box can be dropped (2) 
    \item After stopping box (7), a box is on the robot (21) must start.
     \item After a box is on the robot (21), the second goal started (9) event must start or as an uncontrollable event robot can be failed (4).
     \item After the second goal started (9), second goal reached (11) event must start or as an uncontrollable event robot can be failed (4).
     \item After the second goal reached (11), success flag (13) event must start
     \item After uncontrollable event robot fails (0 and 4), error flag (23) event must start.
     \item After uncontrollable event box dropped (2), spawn box (15) event must start.
	\end{enumerate}

\section{Experiments}
We obtained our DES Supervisor using TCT software and execute autonomous delivery scenario in the physical simulator. 
\subsection{TCT Software}
\label{sec:results}
\begin{figure}
     \includegraphics[width=\linewidth]{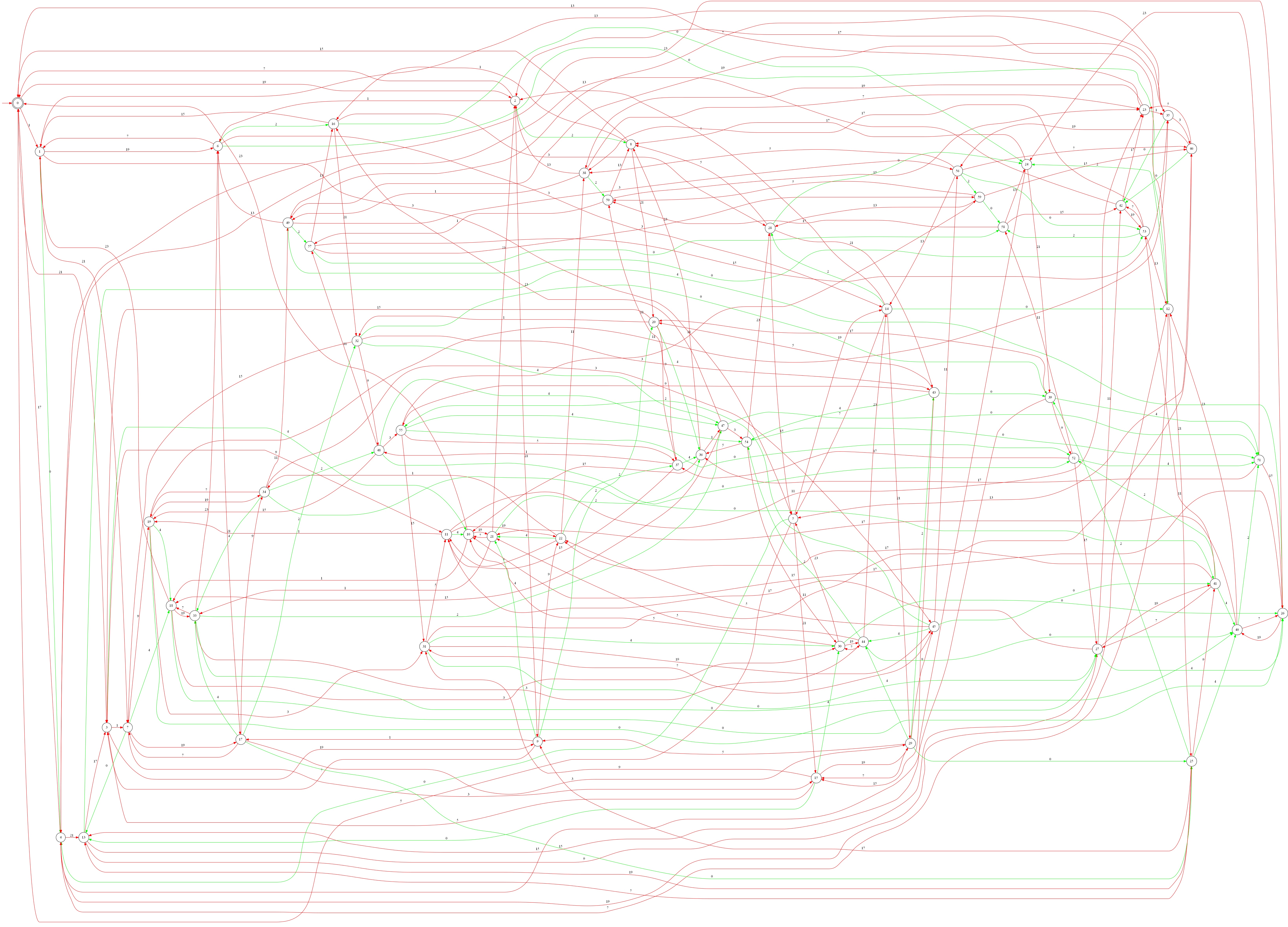}
     \caption{The automaton model of  Free behavior}
     \label{fig:free}
 \end{figure}

This program allows us to the synthesis of supervisory controls for discrete-event systems. In this section, firstly we combined eight specifications into one specification represented as $E$, which is done by using TCT Software's "MEET" function.
Following Figure \ref{fig:specs} can be obtained as a representation of finite-state and events system.
 \begin{figure}
     \includegraphics[width=\linewidth]{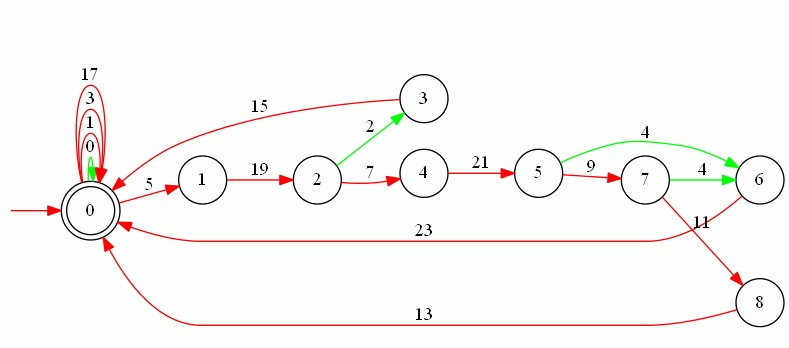}
     \caption{The automaton model of  Specifications: (Red) controllable, (Green) uncontrollable events}
     \label{fig:specs}
 \end{figure}
Secondly, the synchronized combination of Machine-1,Machine-2 and Machine-3 are obtained from using TCT software's "SYNC" function. The results of sequences, which is free behavior of delivery package scenario,  shown in Figure- \ref{fig:free}. As it can be observed that, states and events are massive which includes 60 states and 254 transitions. These states and events contain all the possibility of control sequences which we call free behavior or plant $G$.

  \begin{figure}
     \includegraphics[width=\linewidth]{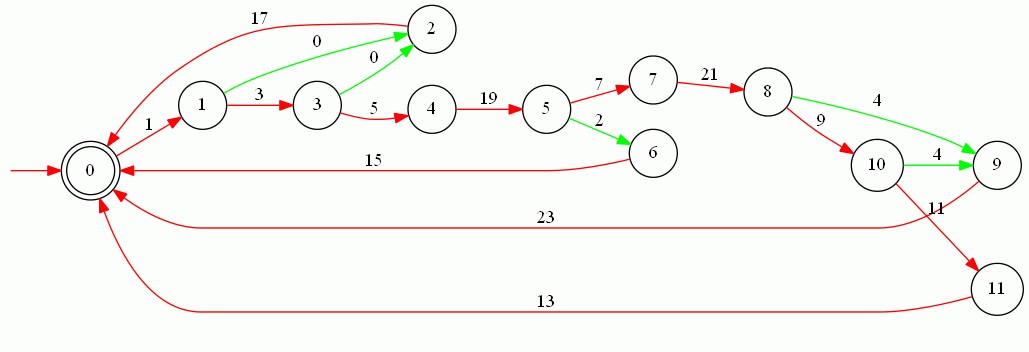}
     \caption{The automaton model of  Controlled behavior: (Red) controllable, (Green) uncontrollable events }
     \label{fig:controlled}
 \end{figure}
Lastly, to obtain controlled behavior, supervisor function ($S$) can be calculated out by using TCT software's "SUPCON" function which uses given specifications ($E$) to control plant ($G$). Thereafter it returns a non-blocking, minimally restrictive supervisor $S/G$.
Controlled behavior of the autonomous delivery package scenario can be shown Figure \ref{fig:controlled}. 
As one of the controlled behavior's states sequence (0-1-3-4-5-7-8-10-11) can confirm that, our controlled behavior meets with specifications and perform its task accordingly.

\subsection{ROS-Gazebo Simulator}
\begin{figure}
     \includegraphics[width=0.49\linewidth]{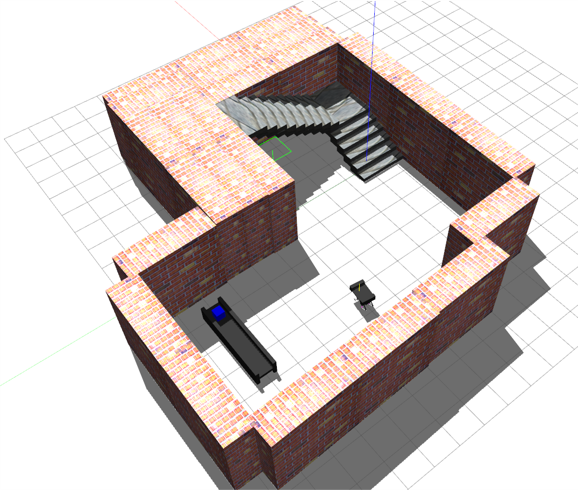}
     \includegraphics[width=0.49\linewidth]{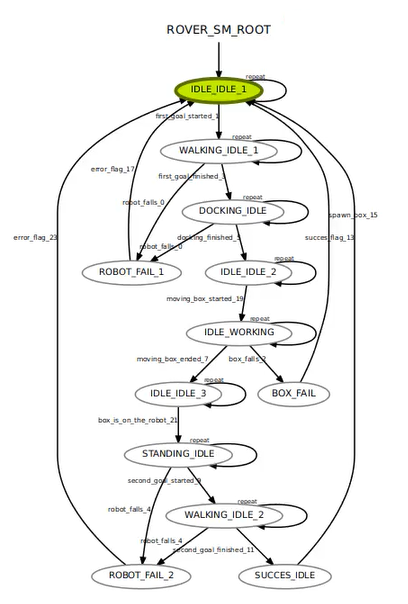}
     \includegraphics[width=0.49\linewidth]{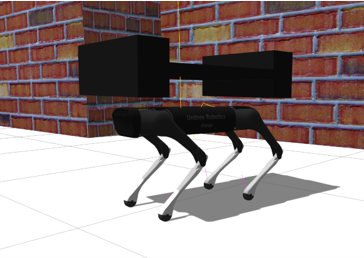}
     \includegraphics[width=0.49\linewidth]{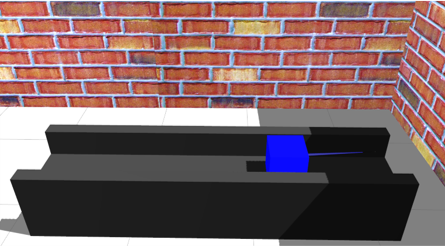}
     \caption{Simulation Environment in Gazebo: (Up-Left) indoor-uneven environment which includes stair, (Up-Right) states and events in ROS-Smach implementation (Bottom Left) Quadrupedal robot which package carrier is mounted it’s back,  (Bottom Right) Conveyor belt which moves box to robot’s back}
     \label{fig:simulation}
 \end{figure}

\begin{figure}
     \includegraphics[width=0.49\linewidth]{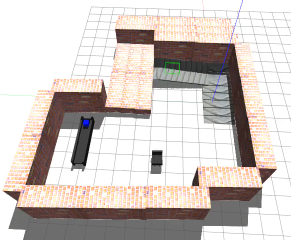}
     \includegraphics[width=0.49\linewidth]{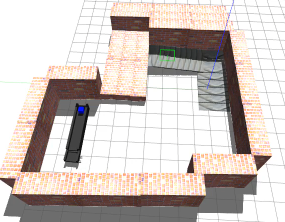}
     \includegraphics[width=0.49\linewidth]{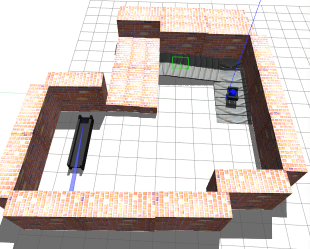}
    \includegraphics[width=0.49\linewidth]{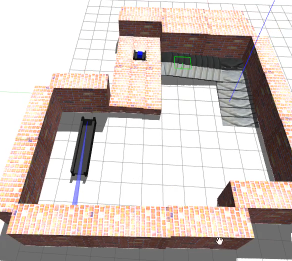}

     \caption{Snapshot of Autonomous Package Delivery Scenario:(Up-Right) initial Position, (Up-Left) first goal reached, (Bottom-Left) climbing to the stairs, (Bottom-Right) second goal reached }
     \label{fig:action}
 \end{figure}

 We have modelled our autonomous package delivery scenario in ROS-Gazebo environment. Our simulation environment similarly includes high walls and stairs which is shown Figure \ref{fig:simulation}-up-left. 
 The controlled behaviour's DES  implemented to ROS-Smach and visualized as Figure \ref{fig:simulation}-up-right.
 Quadrupedal robot is represented as Figure \ref{fig:simulation}-bottom-left and conveyor belt is shown as Figure \ref{fig:simulation}-bottom-right. 
Execution of the state machine based on Supervisory Control Theory is shown Figure \ref{fig:action}. Right to left snapshots shows the scenario of package delivery system in designed world.
Furthermore, we used ROS-Smach library to represent controlled behaviour in the aspect of finite-state and event system.  
Leg Locomotion, path planning and smach algorithms are worked separately in one ROS environment.  
All events are subscribed from the ROS environment to perform actions and all events published to the ROS environment to change its current state. \subsection{Result} 

In the result, we obtained controlled DES behaviour of autonomous package delivery scenario and fully functional simulation environment where a conveyor belt moves box and legged robot navigate through desired goal positions. It was observed that by implementing controlled behaviour of finite-state and events, a robot can take a package from conveyor and deliver-climb to specific location where is one floor above.

\section{Conclusion}
\label{sec:conclude}
In this paper, supervisory control of DES is implemented to analyze the
acceptable control sequence among free sequence for autonomous package delivery scenario using a legged robot and a conveyor belt.
After the defining DES for each machine and their specification, automated model of controlled behaviour is obtained. 
Thereafter, to show the effectiveness of supervisory control theory,  DES of controlled behaviour is implemented on the simulation environment and it successfully performs autonomous package delivery in multi-storey terrain. 
For future work, we can obtain controlled DES model of multiple legged robots and conveyors and test in simulation and the real world.

\end{document}